\documentclass[runningheads]{llncs}

\usepackage{eccv}

\usepackage{eccvabbrv}

\usepackage{graphicx}
\usepackage{booktabs}
\usepackage{multirow}
\usepackage{amssymb}
\usepackage[accsupp]{axessibility}  

\usepackage{floatrow}

\usepackage{hyperref}

\usepackage{orcidlink}

\begin{document}


\title{PanoVOS: Bridging Non-panoramic and Panoramic Views with Transformer \\for Video Segmentation}

\titlerunning{PanoVOS: Bridging Non-panoramic and
Panoramic Views }

\author{Shilin Yan\inst{1} \and
Xiaohao Xu\inst{2} \and
Renrui Zhang\inst{3} \and
Lingyi Hong\inst{1} \and \\
Wenchao Chen\inst{1} \and
Wenqiang Zhang\inst{1} \and
Wei Zhang\inst{1}\thanks{Corresponding author.}
}

\authorrunning{Yan et al.}

\institute{Shanghai Key Lab of Intelligent Information Processing, \\
School of Computer Science, Fudan University \and
University of Michigan, Ann Arbor  \and
MMLab CUHK \and
\email{tattoo.ysl@gmail.com} \quad \email{weizh@fudan.edu.cn}
}

\maketitle
\includegraphics[width=0.95\textwidth]{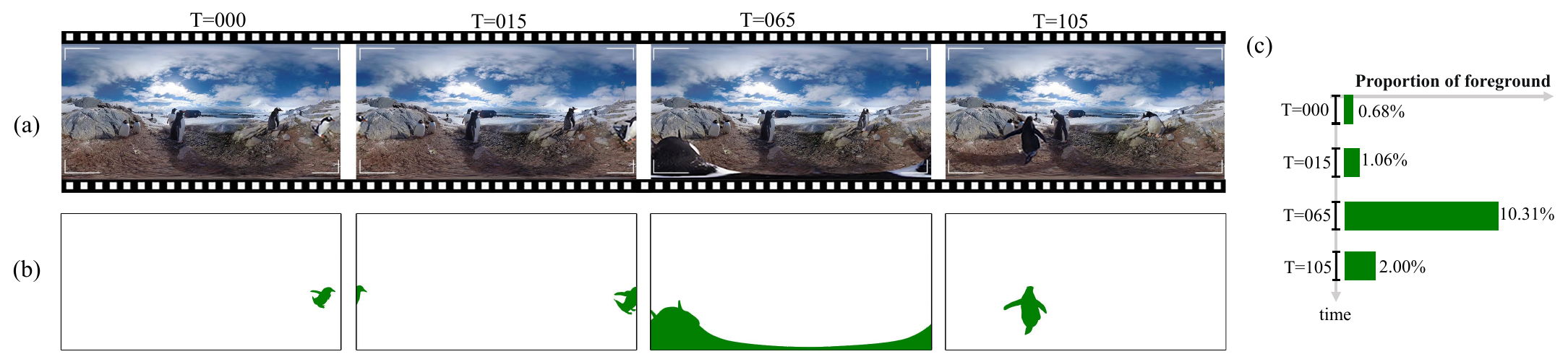} 
\captionof{figure}{\textbf{Panoramic video object segmentation (PanoVOS).} PanoVOS targets tracking and distinguishing the particular instances under content discontinuities (\eg penguin in the image of $T=15$) and serve distortion (\eg penguin in the image of $T=65$). We show the sample of  (a) frames, (b) segmentation annotations, and (c) area proportion of foreground for the \textit{Penguin} video  in our dataset.}
\label{fig:intro}

\begin{abstract}
Panoramic videos contain richer spatial information and have attracted tremendous amounts of attention due to their exceptional experience in some fields such as autonomous driving and virtual reality. However, existing datasets for video segmentation only focus on conventional planar images. To address the challenge, in this paper, we present a panoramic video dataset, \textit{i.e.}, PanoVOS. The dataset provides 150 videos with high video resolutions and diverse motions. To quantify the domain gap between 2D planar videos and panoramic videos, we evaluate 15 off-the-shelf video object segmentation (VOS) models on PanoVOS. Through error analysis, we found that all of them fail to tackle pixel-level content discontinues of panoramic videos. Thus, we present a Panoramic Space Consistency Transformer (PSCFormer), which can effectively utilize the semantic boundary information of the previous frame for pixel-level matching with the current frame. Extensive experiments demonstrate that compared with the previous SOTA models, our PSCFormer network exhibits a great advantage in terms of segmentation results under the panoramic setting. Our dataset poses new challenges in panoramic VOS and we hope that our PanoVOS can advance the development of panoramic segmentation/tracking. The dataset, codes, and pre-train models will be published at~\href{https://github.com/shilinyan99/PanoVOS}{https://github.com/shilinyan99/PanoVOS}.
\end{abstract}

\section{Introduction}
\label{sec:intro}

Semi-supervised video object segmentation (VOS)~\cite{xu2018youtube}, which targets tracking and distinguishing the particular instances across the entire video sequence based on the first frame masks, plays an essential role in video understanding and editing. Conventionally, the images or videos studied in VOS are 2D planar data with a limited Field of View (FoV), which may lead to some ambiguities, especially when objects are out of view. Meanwhile, with the rapid development of VR/AR collection devices~\cite{jost2021quantitative, eger2020measuring}, panoramic videos with a 360$^{\circ}$ $\times$ 180$^{\circ}$ FoV are able to collect the entire viewing sphere and richer spatial information~\cite{ai2022deep, jiang2021cubemap, yuan2021360, li2022panoramic}. To the best of our knowledge, we are the first to attempt to tackle the promising but challenging task of panoramic video object segmentation.

To foster the development of panoramic VOS, we propose a new dataset in this work, aiming at panoramic video object segmentation. The dataset contains a wide range of real-world scenarios in which scenes have a large magnitude of motion. 
The main characteristics of our dataset are three aspects. 
1)  Panoramic videos bring certain advantages (richer geometric information and wider FoV) in real-world applications as well as challenges (serve distortion and content discontinuities). 
2)  Compared to all existing VOS datasets, our dataset has longer video clips with an average length of 20 seconds.
3) Nearly half of the video resolutions in our dataset are $4K$, which may help facilitate broader video tracking/segmentation research under the high-resolution scenario.

{In the proposed dataset}, we annotated 150 videos with 19,145 annotated instance masks, including sports (\eg parkour, skateboard), animals (\eg elephant, monkey), and common objects (basketball, hot balloon). 
{Since, annotating} a pixel-level intensive task is very time-consuming and expensive{, we} proposed a semi-supervised human-computer joint annotation strategy. Concretely, we first annotated {objects} at selected keyframes (1 fps) 
Then we adopted the state-of-the-art video object segmentation model AOT~\cite{yang2021associating} for mask propagation {to the rest frames of videos} and 
{we manually refine parts of them.}

Then, we conduct{ed} extensive experiments on PanoVOS to evaluate 15 off-the-shelf video object segmentation models. The results suggest that existing approaches can not handle several domain-unique challenges. The first is content discontinuities, which means the foreground object may be separated in the left and the right boundaries of the planar image, such as the case in the image of $T=15$ in Fig.~\ref{fig:intro}. The second is the severe distortions and deformations, such as the case in the image of $T=65$ in Fig.~\ref{fig:intro}.

{To tackle these challenges of panoramic video segmentation,}
we proposed a PSCFormer model which consists of key component Panoramic Space Consistent (PSC) blocks. The PSC block is designed for constructing spatial-temporal class-agnostic correspondence and propagating the segmentation masks. Each PSC block utilizes a cross-attention for matching with references' embeddings and a PSC-attention for modeling the boundary semantic relationship between the previous frame and the query frame. Hence, the network can effectively alleviate the problem that the left and the right boundaries are actually continuous in panoramic videos. 
{Our method outperforms the SOTA models that are re-trained on PanoVOS train set in segmentation quality under the panoramic setting.}

Our contributions are three-fold.

\begin{itemize}
	\item We introduce a panoramic video object segmentation dataset {(PanoVOS)} with 150 videos and 19K annotated instance masks, which fills the gap of long-term instance-level annotated panoramic video segmentation datasets.
	\item Extensive experiments are conducted on 15 off-the-shelf VOS benchmarks and our baseline model on PanoVOS, which reveals that current methods could not tackle content discontinuities in panoramic videos well.
	\item We propose a Panoramic Space Consistency Transformer (PSCFormer) on PanoVOS that successfully resolves the {challenges of} discontinuity of pixel-level content {segmentation}.
\end{itemize}

\section{Related Work}
\label{sec:related_work}
\subsection{Panoramic Datasets}

In this paper, \textit{panoramic videos} refers to {complete (360°, full view)} panoramic videos, which is different from the definition in~\cite{mei2022waymo}, which only include {wide but partial views} of some {range-view images} collected from multiple cameras. 

\textbf{Image-based panoramic datasets.} Existing popular image-level panoramic segmentation datasets {are}
Stanford2D3D~\cite{armeni2017joint} and DensePASS~\cite{ma2021densepass}. The former one is mainly focused on indoor spaces including a total of 1,413 panoramic images with instance-level annotations in 13 categories. The latter targets  {driving scenes in cities.} DensePASS~\cite{ma2021densepass} provides only 100 labeled panoramic images for testing and 2,000 unlabeled images for cross-domain transfer optimization. 

\textbf{Video-based panoramic datasets.} Video-based benchmarks mainly include SHD360~\cite{zhang2021shd360}, SOD360~\cite{zhang2018saliency} and Wild360~\cite{cheng2018cube}. All of them are used for panoramic video saliency object detection. Specifically, 
{1) SHD360 only targets human-centric video scenes with little movement.} 
It provides 6,268 object-level pixel-wise masks and 16,238 instance-level pixel-wise masks. 2) SOD360 focuses on the sports-centric scenario with 41 video clips (12 outdoor {and} 29 indoor). 3) Wild360 concentrates on natural 
{scenes with 85 videos.}
{Note that}
SOD360 and Wild360 have no object-level or instance-level annotations.

\begin{table}[t]
\caption{\textbf{Comparison of panoramic video datasets.} Our PanoVOS is the first long-term panoramic video segmentation dataset with instance-level masks. Compared with existing panoramic video datasets~\cite{zhang2021shd360,zhang2018saliency,cheng2018cube} that are used for saliency detection, our panoramic video dataset for video segmentation, \textit{i.e.}, PanoVOS, includes more diverse and larger motion, making it suitable for dense video tracking evaluation.}
	\centering
	\resizebox{0.75\textwidth}{!}
	{
		\begin{tabular}{lccccc}
			\toprule
		\multirow{2}{*}{\textbf{Datasets}}	&  \multirow{2}{*}{\textbf{Motion}}& \multirow{2}{*}{\textbf{\#Videos}}   &  \multirow{2}{*}{\textbf{\#Frames}}& \textbf{\#Total} & \textbf{Average} \\
			 & & && \textbf{Masks}& \textbf{Duration}   \\ \midrule
			SHD360~\cite{zhang2021shd360} & Small & 41 &6,268 & 16,238& 5$s$ \\
			SOD360~\cite{zhang2018saliency} & Large & 104 & N/A & 0 & N/A \\
			Wild360~\cite{cheng2018cube} & Large & 85 & N/A  & 0 & N/A \\ \bottomrule
			PanoVOS                     & Large & \textbf{150} &\textbf{13,995} & \textbf{19,145} & \textbf{20}$s$ \\
			\bottomrule
		\end{tabular}
	} 
	
    \label{table:comparison_datasets}
\end{table}

We make a comparison with the existing video panoramic datasets in Table~\ref{table:comparison_datasets}.
Specifically, our PanoVOS dataset contains 150 videos mainly from three different domains{: person, animal, and common object, which makes the dataset more general for object-agnostic evaluations. 
Besides, videos in our dataset have a relatively large range of motion, making our PanoVOS dataset suitable for video tracking and segmentation evaluation {tasks} under panoramic scenes. Moreover, the average duration of each video in our dataset is 20$s$, which is about 4 times longer than SHD360~\cite{zhang2021shd360} (5$s$ per video). By the way, the longer video is highlighted in a recent survey~\cite{wang2021survey}. The longer the video, the more likely it is to introduce more panoramic video characteristics such as distortion and discontinuity, which is more challenging and more practical.

\subsection{Video Object Segmentation Datasets}
The establishment of DAVIS~\cite{perazzi2016benchmark,pont20172017}and YouTube-VOS~\cite{xu2018youtube} datasets pave the way for the boosting development of VOS methods. They are collected by traditional pinhole cameras and the duration of each video clip is very short, only 5$s$ on average. 
In contrast, the average video length in the proposed PanoVOS dataset is 20s, which is 4 times longer than the existing video datasets.
Our dataset includes more challenging scenes (\eg distortion and discontinuity) that is non-negligible in real-world applications.
}

\subsection{Video Object Segmentation Methods}
Existing video object segmentation methods can be roughly classified into three subsets: online-learning-based, propagation-based, and matching-based. \\
\textbf{Online learning-based.} Online learning-based approaches~\cite{caelles2017one, xiao2018monet,maninis2018video}, which either train or fine-tune their networks with the first-frame ground truth at test time and are therefore a great waste of resources.
OnAVOS~\cite{voigtlaender2017online} achieves promising results by introducing an online adaptation mechanism, but it still requires online fine-tuning. 
To a certain extent, it restricts networks’ efficiency. \\
\textbf{Propagation-based.} Propagation-based models~\cite{chen2020state, cheng2018fast, oh2018fast} get the target masks in a frame-to-frame prorogation way. Although propagation-based methods improve efficiency, they lack long-term context and therefore are difficult to handle object disappearance and reappearance, severe obscuration, and distortion. \\
\textbf{Matching-based.} Matching-based methods~\cite{oh2019video, cheng2021rethinking,zhang2020transductive, mao2021joint,liang2021video,li2024hfvos,li2024onevos, guo2022adaptive,guo2024clickvos,xu2022reliable,xu2022towards,dang2024adaptive,dang2023unified} aim to learn an embedding space of target objects between query and memory.
Recently state-of-the-art methods encode many frames into embeddings and store them as a feature memory bank.
The most representative is STM~\cite{oh2019video}, which has been extended to many works~\cite{seong2020kernelized, hu2021learning, xie2021efficient, wang2021swiftnet,liu2022global,cheng2022xmem}. 
AOT~\cite{yang2021associating} introduces an identification mechanism by encoding multiple targets into the same embedding space, which can simultaneously segment multiple objects. However, they fail to address the challenges of the tremendous proportion of distortion and discontinuity under panoramic  setting.

\begin{figure*}[t]
	\centering
	\includegraphics[width=\textwidth]{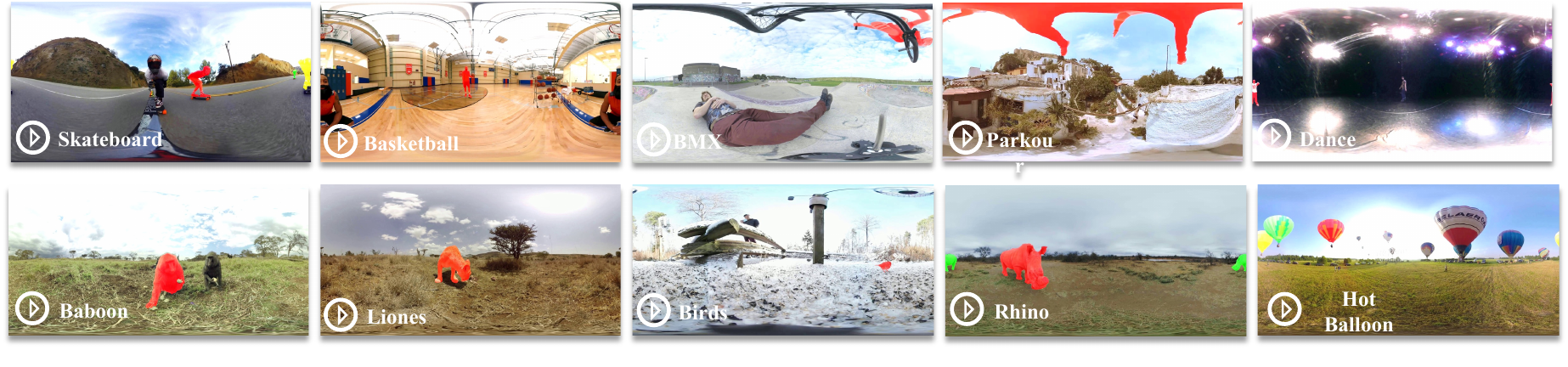} 
	\caption{\textbf{PanoVOS dataset.} We select 10 samples from the dataset involving major scenes. For each video, there are high-quality instance-level pixel-wise masks.}
	\label{fig:dataset_show}
\end{figure*}

\begin{table}[t]
	\caption{{\textbf{Statistics of PanoVOS dataset}}}
	\centering

	\resizebox{0.6\textwidth}{!}
	{
		\begin{tabular}{lccccc}
			\toprule
			\textbf{Splits} & \textbf{Train} & \textbf{Val} & \textbf{Test} \\ \midrule
			\#Videos & 80 & 35 (10 unseen) & 35 (10 unseen)  \\
			\#Images & 7,070 (50.5$\%$) & 3,464 (24.8$\%$)   & 3,461 (24.7$\%$) \\
			\#Masks & 9,585 (50.1$\%$) & 4,957  (25.9$\%$)  & 4,603 (24.0$\%$)  \\ 
			\bottomrule
		\end{tabular}
	} 
	
	\label{table:PanoVOS_statistics}
\end{table}

\section{PanoVOS Dataset}
{We introduced the proposed PanoVOS dataset in three parts, (1) collection process, (2) statistical summary, and (3) annotation pipeline. }

\subsection{Data collection} We built our PanoVOS dataset with the principle of diversity in mind. Moreover, the objects in the video should have a large amplitude of motion or camera movement. {Based on the above viewpoint, we collected videos from the YouTube website for further annotation, respectively.} 
{The range of the video length is from 3 to 40 seconds.}
The average sequence length of each video in the dataset is approximately 20 {seconds}. We {followed the settings of YouTube-VOS~\cite{xu2018youtube} to sample} the frames at 6 fps.

 \begin{figure*}[t!]
	\centering
	\includegraphics[width=0.85\textwidth]{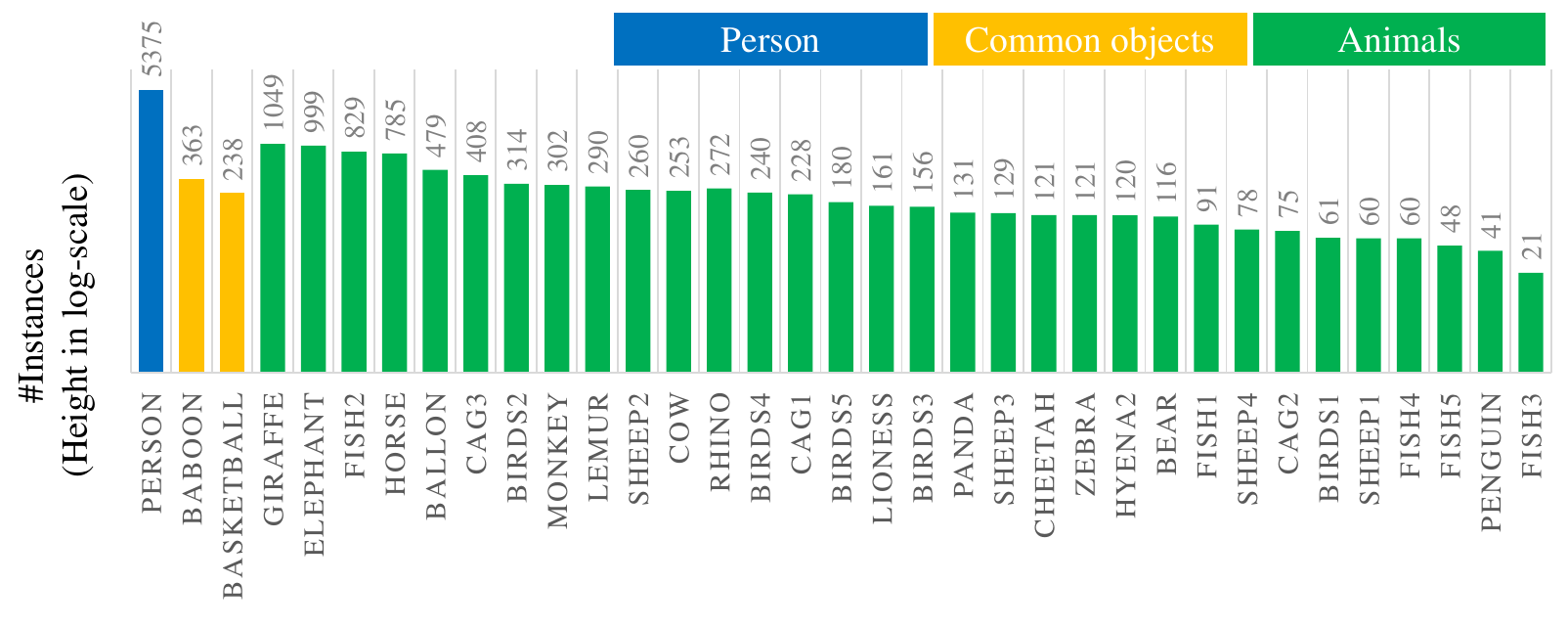} 
 \caption{\textbf{{Instance-level distribution of PanoVOS dataset.}} Our dataset contains three major divisions: \textit{person}, \textit{animals}, and\textit{ common objects} with \textcolor{black}{35} sub-divisions.}
	\label{fig:dataset_cag}
\end{figure*}

\subsection{Dataset Statistics}
PanoVOS contains 150 videos, including 13,995 frames and 19,145 instance annotations from 35 categories. The average length of each video is 20 seconds. We believe that visual categories are representative of common life scenarios, and Fig.\ref{fig:dataset_show} shows some samples of PanoVOS. To create our PanoVOS, in the spirit of the video object segmentation task, we carefully selected videos with relatively large motion amplitudes and chose a set of video categories including person (\emph{e.g.} parkour, dance, BMX, skateboard), animals (\emph{e.g.} elephant, monkey, giraffe, rhino, birds) and common objects (\emph{e.g.} basketball, hot balloon) as shown in Fig~\ref{fig:dataset_cag}. 
 PanoVOS dataset consists of 150 videos split into training (80), validation (35), and test (35) sets. Table~\ref{table:PanoVOS_statistics} shows detailed division results. Both the validation and test sets have 35 videos (about 23$\%$ of the frames and the masks). For validation and test sets, we keep some unseen visual categories for generalization ability evaluation.

\subsection{Annotation Pipeline}
\begin{figure*}[t]
	\centering
	\includegraphics[width=1\textwidth]{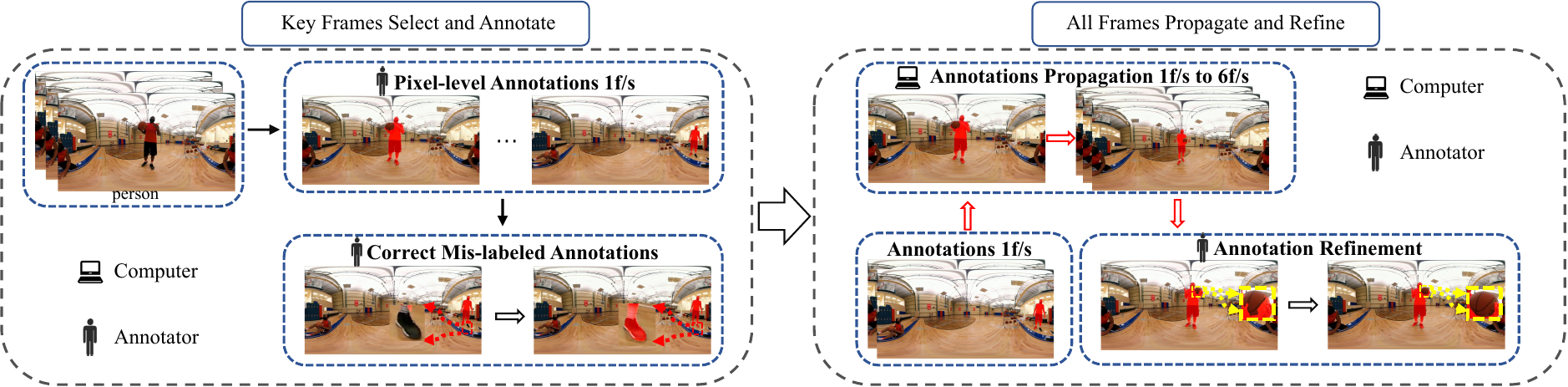} 
	\caption{\textbf{PanoVOS annotation pipeline.} Our annotation pipeline includes two phases. (1) The first phase is called \textit{Key Frames Select and Annotate}. The annotator browses the video and picks out the object to be annotated. Then, instances are manually annotated at 1 fps and corrected by another annotator. (2) The second phase is called \textit{All Frames Propagate and Refine}. In this phase, we apply a semi-supervised video object segmentation model to help propagate the annotated masks and the generated instances are refined by annotators.}

	\label{fig:annotation_pipeline}
\end{figure*}

Annotation is very time-consuming and expensive for a pixel-level panoramic segmentation dataset. 
To obtain accurate large-scale video panoramic segmentation annotations and make the process more efficient, we propose a semi-automatic human-computer joint annotation strategy, as shown in Fig~\ref{fig:annotation_pipeline}. First, keyframes are selected and manually annotated for each video, which are images with a speed of 1 fps. This is followed by a frame-by-frame propagation from the annotated keyframes to those unlabeled intermediate frames with a sophisticated semi-supervised VOS model. Then, to tackle the distortions and discontinuities in panoramic videos, we need to re-calibrate the resulting annotations via human refinement. More details will unfold below.

\subsubsection{Annotation Propagation} 
For the annotation of each video, we first need an expert to browse the current video and note down all objects that have a large amplitude of movement. Then, for each video, the recorded objects in keyframes with a speed of 1 fps are selected for manual annotation. To avoid consistency errors or the problem of objects being labeled as other instances when they disappear and reappear, another expert needs to double-check the annotations of all objects to improve the accuracy of the dataset annotation.

We then use the off-the-shelf video labeling method ~\cite{yang2021associating} to propagate the instance masks frame by frame from the annotated keyframes to untagged intermediate frames and generate masks at 6 fps.

\subsubsection{Annotation Refinement}
{To present a new Panoramic dataset of high quality.}
After obtaining masks of the first propagation stage, annotators are asked to check the quality of the masks and refine them. The main amendments are in the following two areas. 1). Since our video resolution is generally relatively high, the propagation method will often fail when encountering complex videos with many small objects in a scene. 2) Due to the huge distortions and discontinuities present in the panoramic video, the quality of the masks obtained is relatively poor. Manual correction of the mask is checked by another annotator until the result is satisfactory before proceeding to the next video annotation.

\section{Method}

\subsection{Overview}

Video object segmentation targets assigning an instance label to every pixel in the given video sequence based on the first frame mask. Recent works~\cite{yang2021collaborative, cheng2021rethinking, cheng2022xmem, fang2023instructseq} have demonstrated that the attention mechanism can significantly help improve the segmentation performance. However, for the challenge of content discontinuation in panoramic videos, only considering the original attention mechanism will not be able to fully utilize the semantic information on the left and right boundaries (pixel contiguity) in the spatial dimension and will lose valuable contextual information when segmenting objects. Therefore, in this work, our mission is to design an effective network architecture, which can help acquire valuable boundary relationships.

\begin{figure*}[t]
	\centering
	\includegraphics[width=0.85\textwidth]{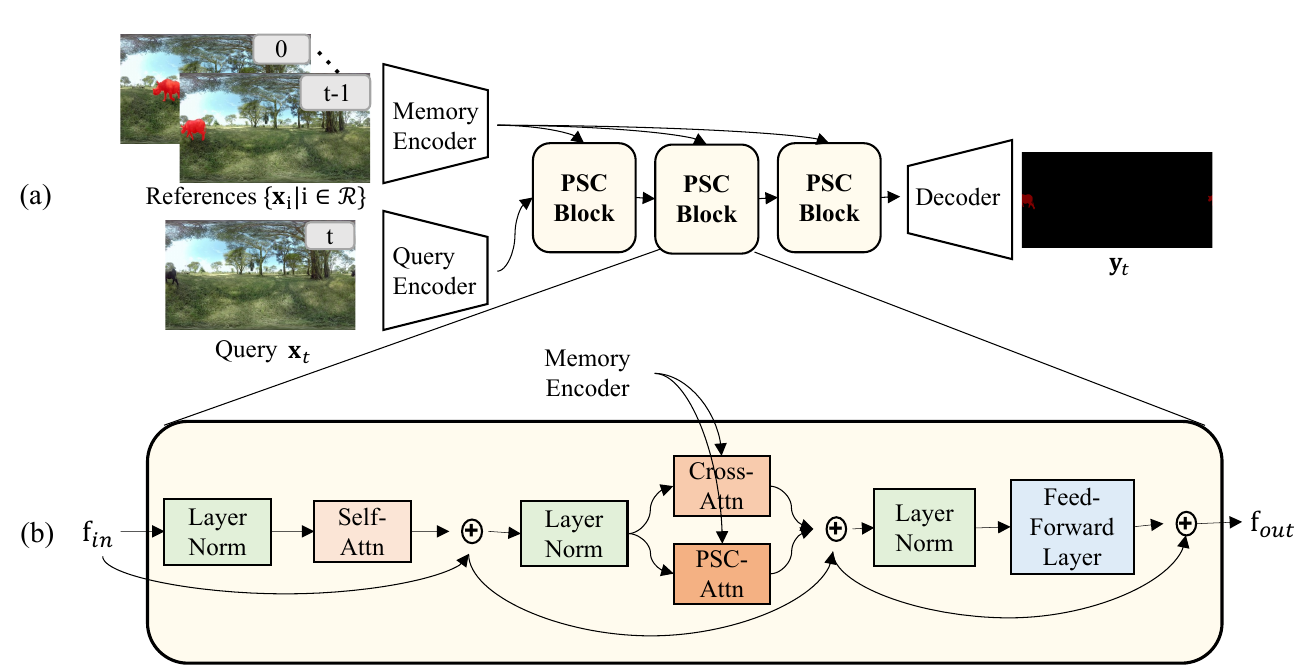} 
	\caption{(a) \textbf{PSCFormer overview.} Given the query frame $\mathbf{x}_t$ and reference frames $\{\mathbf{x}_i| i\in\mathcal{R} \}$, the goal of VOS is to delineate objects from the background by generating mask $\mathbf{y}_t$ for query frame $\mathbf{x}_t$. References and the query frame are encoded by the memory encoder and query encoder, respectively. Multiple stacking panoramic space consistency (PSC) blocks are used to leverage the correspondence in the panoramic space between references and the query frame. A decoder is used for generating the prediction of the query frame. 
	(b) \textbf{Panoramic space consistency block architecture details.} 
 }
	\label{fig:pipeline}
\end{figure*}

Fig.~\ref{fig:pipeline}(a) illustrates the overall architecture of the proposed network. Given the query frame $\mathbf{x}_t$ and references $\{\mathbf{x}_i| i\in\mathcal{R} \}$, the goal of VOS is to delineate objects from the background by generating mask $\mathbf{y}_t$ for query frame $\mathbf{x}_t$. Following~\cite{yang2020collaborative}, our basic setting uses the first and previous frame as references $\mathcal{R}=\{1, t-1\}$. The memory encoder and query encoder are responsible for extracting frame-level features. After this, the panoramic space consistency block takes them as input and aggregates the spatial-temporal information between the reference frames and the query frame at the pixel level. Finally, the decoder uses the output of the sequence stacking PSC blocks to predict the mask of the object.

\subsection{Panoramic Space Consistency Block}

Fig.~\ref{fig:pipeline}(b) shows the structure of a PSC block. Motivated by the common transformer blocks~\cite{vaswani2017attention}, PSC firstly contains a self-attention layer, which is used to aggregate the target objects’ correlation information within the query frame. Then, the middle module is composed of cross-attention and PSC-attention, in which cross-attention is responsible for learning the target objects’ information from references $\mathcal{R}$ and the PSC-attention targets on exploring the boundary relationship between the query frame and previous frame. Finally, PSC employs a two-layer feed-forward MLP with GELU~\cite{hendrycks2016gaussian} non-linearity activation function. 

\textbf{Panoramic Space Consistency Attention (PSC-Attn)}. PSC-Attn is employed to model the spatial-temporal relationship between the query frame and reference frames considering the continuity of pixels of images in the panoramic space. How to establish a connection between the left and right boundaries become especially important? The most intuitive solution would be to directly splice in length, but this would lead to a huge amount of computation. Therefore, we take the approach of moving a portion of the region in the length dimension from the right boundary to the leftmost boundary for stitching. Consequently, we only focus on the left and right boundaries between the query frame and the reference frame. Thus, unlike the original attention, where each query token is counted for attention along with all key tokens in the reference frame, our PSC attention takes care of the key tokens in a fixed window size.
In particular, we define the reference frame feature embedding $\mathbf{f}(\mathbf{x})  \in \mathbb{R}^{H \times W \times C} $, which is extracted from the query encoder. $H$, $W$, and  $C$ represent the height, width, and channel dimensions, respectively. 
According to the solutions mentioned above, the new feature embedding $\mathbf{f(x)}^{\prime}$ is calculated as follows:
\begin{equation}
	\begin{small}
		\begin{aligned}
			\mathbf{f}(\mathbf{x})^{\prime}\left[0: W / p\right] &=\mathbf{f}(\mathbf{x})\left[ W / p: W\right]  \\
			\mathbf{f}(\mathbf{x})^{\prime}\left[W / p: W\right] &=\mathbf{f}(\mathbf{x})\left[ 0: W / p\right]   \\
			\mathbf{f}(\mathbf{x})^{\prime}\left[W / p: W - W / p\right] &=\mathbf{f}(\mathbf{x})\left[ W / p : W - W / p\right],
		\end{aligned}
	\end{small}
\end{equation}
where $p \in \mathbb{Z^+}$.  We define query embedding $Q  \in \mathbb{R}^{HW \times C} $, key embedding $K \in \mathbb{R}^{HW \times C} $, value embedding $V \in \mathbb{R}^{HW \times C}$, where $Q$ is from the query frame feature embedding, $K$ and  $V$ are from $\mathbf{f(x)}^{\prime}$ by performing dimensional transformations. Mathematically, we define the PSC attention as follows,
\begin{equation}
	\operatorname{PSCAttn}(Q, K, V)=\operatorname{softmax}\left(\frac{Q K^{T} \mathbf{R}}{\sqrt{C}}\right) V,
\end{equation}
where $\mathbf{R} \in \left[0, 1\right]^{HW \times HW}$ means a window that represents the attention range of each query token. For query $Q_{(x,y)}$ at $(x, y)$ position, we define the $\mathbf{R}_{(x,y)}$ as:
\begin{equation}
	\mathbf{R}_{x, y}(i, j)= \begin{cases}1 & \text { if }(x-i)^2 \leqslant s^2 \text { and }(y - j)^2 \leqslant s^2 \\
		0 & \text { otherwise }\end{cases},
\end{equation}
where $(i, j)$ is the position for each key token, $s$ is the window size. For each query token, it calculates the attention with another key token only if they are spatially limited to a $(2 \times s + 1)$ size window, which significantly reduces the time complexity from $(h \times w)^2$ to $(2 \times s + 1)^2$.

\renewcommand{\arraystretch}{0.5}
\begin{table*}[t]
\caption{\textbf{Domain transfer result of (static image datasets)$\to$(PanoVOS Validation \& Test).} Subscript $s$ and $u$ denote scores in seen and unseen categories. $MF$ denotes  multiple historical frames as reference.  \textcolor{blue}{$\downarrow$} represents the performance of the declining values compared to the  YouTube-VOS dataset~\cite{xu2018youtube}. ${*}$ denotes a large-scale external dataset BL30K~\cite{cheng2021rethinking} dataset is used during training.}
	\centering
	\setlength{\tabcolsep}{2.6mm}
	\resizebox{\textwidth}{!}
	{
			\begin{tabular}{l|c|c|lllll|lllll}
			\toprule
			\multirow{2}{*}{Methods}& & YouTube-VOS & \multicolumn{5}{c|}{PanoVOS Validation}  & \multicolumn{5}{c}{PanoVOS Test}                \\
			\cmidrule(l){4-13} &$MF$  & \textbf{$\mathcal{J}\&\mathcal{F}$}
			& \textbf{$\mathcal{J}\&\mathcal{F}$} & $\mathcal{J}_{s}$ & $\mathcal{F}_{s}$ & $\mathcal{J}_{u}$ & $\mathcal{F}_{u}$ & \textbf{$\mathcal{J}\&\mathcal{F}$} & $\mathcal{J}_{s}$ & $\mathcal{F}_{s}$ & $\mathcal{J}_{u}$ & $\mathcal{F}_{u}$\\
			\midrule 
			AOTT {\cite{yang2021associating}}&  &73.7 &53.8\textcolor{blue}{$\downarrow_{19.9}$}& 44.4 & 58.3 & {46.9} & {65.7}  & 43.7\textcolor{blue}{$\downarrow_{30.0}$} & {36.3} & {49.8} & 39.6 & 49.2 \\
			AOTS {\cite{yang2021associating}}&  &74.6 &55.8 \textcolor{blue}{$\downarrow_{18.8}$}& 49.1 & {62.2} & 46.4 & 65.5  & {44.7}\textcolor{blue}{$\downarrow_{29.9}$} & 32.4 & 43.3 & {46.5} & {56.8} \\
			AOTB {\cite{yang2021associating}}&  &{75.2} &53.7\textcolor{blue}{$\downarrow_{21.5}$}& 46.2 & 58.1 & 46.3 & 64.1  & 39.5\textcolor{blue}{$\downarrow_{35.7}$} & 34.4 & 44.4 & 35.0 & 44.4 \\
			\midrule
			AFB-URR{\cite{liang2020video}}&\checkmark&65.2 &40.1\textcolor{blue}{$\downarrow_{25.1}$}& 31.1 & 41.5 & 35.8  & 51.8 & 30.4\textcolor{blue}{$\downarrow_{34.8}$} & 23.1 & 32.7 & 28.8  & 36.9\\
			
			STCN {\cite{cheng2021rethinking}}&\checkmark&76.1 &49.9\textcolor{blue}{$\downarrow_{26.2}$}& 42.7 & 53.4 & 45.1 & 58.4  & 48.0\textcolor{blue}{$\downarrow_{28.1}$}& {39.3} & 50.2 & 46.7 & 55.7 \\
			
			XMem {\cite{cheng2022xmem}}&\checkmark&{77.0} &48.6\textcolor{blue}{$\downarrow_{28.4}$}& 40.7 & 50.1 & 44.8 & 58.6  & 40.2\textcolor{blue}{$\downarrow_{36.8}$}& 35.3 & 44.9 & 36.4 & 44.0 \\
		
			AOTL {\cite{yang2021associating}}&\checkmark&74.7  &49.2\textcolor{blue}{$\downarrow_{25.5}$}& 43.3 & 57.3 & 38.9 & 57.1  & 38.2\textcolor{blue}{$\downarrow_{36.5}$} & 32.3 & 43.7 & 32.7 & 44.1 \\
			R50\_AOTL {\cite{yang2021associating}}&\checkmark&76.5  &50.1\textcolor{blue}{$\downarrow_{26.4}$}& 44.5 & 58.6 & 40.3 & 57.2 & 41.4\textcolor{blue}{$\downarrow_{35.1}$}  & 33.7 & 45.0 & 38.3 & 48.4 \\
			SwinB\_AOTL {\cite{yang2021associating}}&\checkmark&74.4  &44.8\textcolor{blue}{$\downarrow_{29.6}$}& 39.1 & 52.2 & 34.9& 53.0 & 36.2\textcolor{blue}{$\downarrow_{38.2}$}  & 31.1 & 42.0 & 31.0 & 40.6  \\
			\midrule
   RDE$^{*}$ {\cite{li2022recurrent}}&\checkmark&61.7 &43.1\textcolor{blue}{$\downarrow_{18.6}$}& 36.0 & 48.4 & 35.2& 52.7  & 41.3\textcolor{blue}{$\downarrow_{20.4}$} & 30.9 & 44.6 & 41.4 & 48.5 \\
          STCN$^{*}$ {\cite{cheng2021rethinking}}&\checkmark&56.3 &43.2\textcolor{blue}{$\downarrow_{13.1}$} & 41.6 & 53.7 & 33.2 & 44.5 & 38.0\textcolor{blue}{$\downarrow_{18.3}$} & 32.8 & 43.2 & 35.5 & 40.4 \\	
           XMem$^{*}$ {\cite{cheng2022xmem}}&\checkmark & 65.8  & {55.9\textcolor{blue}{$\downarrow_{9.9}$}}& {52.2} & {64.0} & {47.2} & {60.0}  & {49.6}\textcolor{blue}{$\downarrow_{16.2}$}& 39.2 & {52.6} & {46.8} & {59.9} \\
			\bottomrule
		\end{tabular}
	}

	\label{table:coco_PanoVOS_results}
\end{table*}

\renewcommand{\arraystretch}{0.5}
\begin{table*}[t]
	\caption{\textbf{Domain transfer result of (static image datasets \& YouTubeVOS)$\to$(PanoVOS Validation \& Test).} Subscript $s$ and $u$ denote scores in seen and unseen categories. $MF$ denotes  multiple historical frames as reference.  \textcolor{blue}{$\downarrow$} represents the performance of the declining values compared to the YouTube-VOS dataset~\cite{xu2018youtube}. ${*}$ denotes a large-scale external dataset BL30K~\cite{cheng2021rethinking} dataset is used during training. ${\dag}$ denotes no synthetic data is used during the training stage. }
	\centering
	\setlength{\tabcolsep}{2.6mm}
	\resizebox{\textwidth}{!}
	{
		\begin{tabular}{l|c|c|lllll|lllll}
			\toprule
	\multirow{2}{*}{Methods}& & YouTube-VOS & \multicolumn{5}{c|}{PanoVOS Validation}  & \multicolumn{5}{c}{PanoVOS Test}                \\
			\cmidrule(l){4-13} &$MF$  & \textbf{$\mathcal{J}\&\mathcal{F}$}
			& \textbf{$\mathcal{J}\&\mathcal{F}$} & $\mathcal{J}_{s}$ & $\mathcal{F}_{s}$ & $\mathcal{J}_{u}$ & $\mathcal{F}_{u}$ & \textbf{$\mathcal{J}\&\mathcal{F}$} & $\mathcal{J}_{s}$ & $\mathcal{F}_{s}$ & $\mathcal{J}_{u}$ & $\mathcal{F}_{u}$\\
			\midrule 
			CFBI$^{\dag}$ {\cite{yang2021collaborative}}& & 81.4&  60.9\textcolor{blue}{$\downarrow_{20.5}$}& 53.0 & 65.2 & 56.3 & 69.0  & 49.0\textcolor{blue}{$\downarrow_{32.4}$} & 49.4 & 47.6 & 46.2 & 52.6 \\
			CFBI+$^{\dag}$ {\cite{yang2021collaborative}}& & 82.8 &  57.6\textcolor{blue}{$\downarrow_{25.2}$}& 52.1 & 67.0 & 48.1 & 63.4  & 53.7\textcolor{blue}{$\downarrow_{29.1}$} & 51.6 & 59.3 & 46.6 & 57.5 \\
			AOTT {\cite{yang2021associating}}& & 80.2&  61.5\textcolor{blue}{$\downarrow_{18.7}$}& 55.6 & 67.7 & 54.6 & 68.2  & 52.6\textcolor{blue}{$\downarrow_{27.6}$}& 44.8 & 55.3 & 51.5 & 58.8 \\
			AOTS {\cite{yang2021associating}}& & 82.6&  66.7\textcolor{blue}{$\downarrow_{15.9}$}& 58.0 & 70.5 & 62.0 & 76.4  & 57.3\textcolor{blue}{$\downarrow_{25.3}$} & 50.2 & 61.0 & 54.6 & 63.5 \\
			AOTB {\cite{yang2021associating}}& & {83.5}&  70.5\textcolor{blue}{$\downarrow_{13.0}$}& 59.2 & 71.7 & {68.5} & {82.7}  & {60.8}\textcolor{blue}{$\downarrow_{22.7}$} & 53.0 & {64.4} & {57.8} & {68.2} \\
			\midrule 
			AFB-URR{\cite{liang2020video}}&\checkmark & 79.6  & 55.1\textcolor{blue}{$\downarrow_{24.5}$}& 44.7 & 55.6 & 53.4  & 66.7  &52.4\textcolor{blue}{$\downarrow_{27.2}$} & 43.6 & 54.2 & 52.0  & 59.9\\
			RDE {\cite{li2022recurrent}}&\checkmark & 81.9 &  54.7\textcolor{blue}{$\downarrow_{27.2}$}& 50.3 & 63.9 & 44.6 & 60.1  & 55.4\textcolor{blue}{$\downarrow_{26.5}$} & 45.5 & 59.2 & 51.0 & 65.9 \\
			STCN {\cite{cheng2021rethinking}}&\checkmark & 83.0 &  61.8\textcolor{blue}{$\downarrow_{21.2}$}& 50.3 & 63.5 & 61.3 & 72.1  & 53.4\textcolor{blue}{$\downarrow_{29.6}$}& 46.2 & 58.9 & 49.0 & 59.9 \\
			
			XMem {\cite{cheng2022xmem}}&\checkmark & 85.7& 66.1\textcolor{blue}{$\downarrow_{19.6}$}& 56.6 & 68.7 & 62.0 & 77.2  & {62.5}\textcolor{blue}{$\downarrow_{23.2}$} & 53.1 & 65.4 & {61.1} & {70.4} \\
			
			AOTL {\cite{yang2021associating}}&\checkmark& 83.8&  71.9\textcolor{blue}{$\downarrow_{11.9}$}& 62.1 & 75.3 & 67.4 & 82.8  & 62.1\textcolor{blue}{$\downarrow_{21.7}$}& 57.1 & {69.0} & 56.2 & 66.1 \\
			R50\_AOTL {\cite{yang2021associating}}&\checkmark& 84.1&  69.2\textcolor{blue}{$\downarrow_{14.9}$}& 56.7 & 69.4 & 67.5 & 83.1 & 61.4\textcolor{blue}{$\downarrow_{22.7}$} & {57.5} & {69.0} & 53.3 & 65.7 \\
			SwinB\_AOTL {\cite{yang2021associating}}&\checkmark& 84.5&  67.5\textcolor{blue}{$\downarrow_{17.0}$}& 60.2 & 73.6 & 60.3& 76.0 & 60.9\textcolor{blue}{$\downarrow_{23.6}$}  & 53.9 & 63.7 & 58.7 & 67.4  \\
            \midrule
   	RDE$^{*}$ {\cite{li2022recurrent}}&\checkmark& 83.3&  60.9\textcolor{blue}{$\downarrow_{22.4}$}& 51.4 & 64.7 & 56.0& 71.6  & 55.6\textcolor{blue}{$\downarrow_{27.7}$} & 48.1 & 60.8 & 52.6 & 61.0 \\         
    STCN$^{*}$ {\cite{cheng2021rethinking}}&\checkmark & 84.3& 61.7\textcolor{blue}{$\downarrow_{22.6}$}& 49.9 & 61.8 & 59.7 & 75.5 & 55.8\textcolor{blue}{$\downarrow_{28.5}$}& 48.2 & 59.8 & 52.7 & 62.5 \\
    XMem$^{*}$ {\cite{cheng2022xmem}}&\checkmark & {86.1}&  63.4\textcolor{blue}{$\downarrow_{22.7}$}& 53.5 & 64.4 & 61.5 & 74.1  & 61.0\textcolor{blue}{$\downarrow_{25.1}$} & 53.5 & 65.1 & 57.5 & 68.0 \\
			\bottomrule
		\end{tabular}
	} 
	
    \label{table:ytb_PanoVOS_results}
\end{table*}

\renewcommand{\arraystretch}{1}

\renewcommand{\arraystretch}{0.5}
\begin{table*}[h]
\caption{\textbf{Quantitative comparison on PanoVOS for variations of foundation model Segment Anything Model~\cite{kirillov2023segment}}. Subscript $s$ and $u$ denote scores in seen and unseen categories.}
	\centering
	\setlength{\tabcolsep}{4mm}
	\resizebox{\textwidth}{!}
	{
		\begin{tabular}{l|lllll|lllll}
			\toprule
			\multirow{2}{*}{Methods} & \multicolumn{5}{c|}{PanoVOS Validation}  & \multicolumn{5}{c}{PanoVOS Test}                \\
			\cmidrule(l){2-11} 
			& $\mathcal{J}\&\mathcal{F}$ & $\mathcal{J}_{s}$ & $\mathcal{F}_{s}$ & $\mathcal{J}_{u}$ & $\mathcal{F}_{u}$ & $\mathcal{J}\&\mathcal{F}$ & $\mathcal{J}_{s}$ & $\mathcal{F}_{s}$ & $\mathcal{J}_{u}$ & $\mathcal{F}_{u}$\\
			\midrule 
			PerSAM~\cite{zhang2023personalize}&19.1 &12.3& 19.8 & 17.4 & 27.1 & 19.5  & 7.4 & 14.9 & 23.8 & 31.7  \\
               SAM-PT~\cite{rajivc2023segment}&47.5 &36.7& 48.6 & 46.0 & 58.7 & 41.0  & 31.1 & 40.5 & 40.2 & 52.3  \\
                SAM-PT-reinit~\cite{rajivc2023segment}&43.7 &34.3& 44.3 & 41.3 & 54.9 & 43.6  & 35.0 & 42.7 & 43.5 & 53.0  \\
			\bottomrule
		\end{tabular}
	} 
    	
    \label{table:sam_PanoVOS_results}
\end{table*}

Following~\cite{vaswani2017attention}, we implement the representational form of our PSCAttn module with multi-headed attention, defined mathematically as follows,
\begin{equation}
	\begin{small}
		\begin{aligned}	\operatorname{MultiHead}(Q, K, V) &=\operatorname{Concat}\left(\text {head}_1, \ldots, \text {head}_h\right) W^O \\
			\text {head}_i &=\operatorname{PSCAttn}\left(Q W_i^Q, K W_i^K, V W_i^V\right),
		\end{aligned}
	\end{small}
\end{equation}
where $W_i^Q \in \mathbb{R}^{C \times d_{model}}$, $W_i^K \in \mathbb{R}^{C \times d_{model}} $ , $W_i^V \in \mathbb{R}^{C \times d_{model}} $ and $W_i^O \in \mathbb{R}^{C \times C}$ are the linear projections.  As~\cite{vaswani2017attention}, we set the number of heads to $\left(h = C / d_{model} \right)$ 8, where $d_{model}$ is the projection dimension of each head.

\renewcommand{\arraystretch}{0.5}
\begin{table*}[h]
\caption{\textbf{Quantitative comparison on PanoVOS for models with pretraining on static image datasets}. Subscript $s$ and $u$ denote scores in seen and unseen categories. $MF$ denotes  multiple historical frames as reference.  ${*}$ denotes a large-scale external dataset BL30K~\cite{cheng2021rethinking} dataset is used during training. }
	\centering
	\setlength{\tabcolsep}{4mm}
	\resizebox{\textwidth}{!}
	{
		\begin{tabular}{l|c|lllll|lllll}
			\toprule
			\multirow{2}{*}{Methods} & & \multicolumn{5}{c|}{PanoVOS Validation}  & \multicolumn{5}{c}{PanoVOS Test}                \\
			\cmidrule(l){3-12} &$MF$
			& $\mathcal{J}\&\mathcal{F}$ & $\mathcal{J}_{s}$ & $\mathcal{F}_{s}$ & $\mathcal{J}_{u}$ & $\mathcal{F}_{u}$ & $\mathcal{J}\&\mathcal{F}$ & $\mathcal{J}_{s}$ & $\mathcal{F}_{s}$ & $\mathcal{J}_{u}$ & $\mathcal{F}_{u}$\\
			\midrule 
			CFBI$^{\dag}$ {\cite{yang2020collaborative}}& & 35.8 & 34.6 & 44.8 & 24.2 & 39.7  & 19.1 & 18.2 & 26.1 & 12.2 & 19.8 \\
			CFBI+$^{\dag}$ {\cite{yang2021collaborative}}& & 41.3 & 38.0 & 47.9 & 32.5 & 46.9  & 30.9 & 30.8 & 42.7 & 21.4 & 28.5 \\
			AOTT {\cite{yang2021associating}}& &  65.6& 59.4 & 68.3 & 59.7 & 75.0  & 53.4 & 49.3 & 61.6 & 47.5 & 55.1 \\
			AOTS {\cite{yang2021associating}}& &  67.7& 61.2 & 70.0 & 62.4 & 77.1  & 55.9 & \textbf{53.2} & \textbf{65.1} & 48.6 & 57.0 \\
			AOTB {\cite{yang2021associating}}& &  67.6& 62.3 & 72.0 & 61.5 & 74.8  & 55.4 & 53.5 & 64.2 & 47.7 & 56.0 \\
			\textbf{Ours-Base}               & & \textbf{74.0}& \textbf{66.4} & \textbf{80.4} & \textbf{66.2} & \textbf{83.0} & \textbf{56.8} & 49.4 & 62.7 & \textbf{52.4} & \textbf{62.5} \\
			\midrule
			AFB-URR{\cite{liang2020video}}&\checkmark& 34.3 & 34.8 & 42.8 & 24.9  & 34.5 & 34.2 & 28.2 & 38.8 & 32.9  & 36.8\\
			RDE {\cite{li2022recurrent}}&\checkmark& 50.5 & 49.7 & 58.4 & 39.2 & 54.9  & 42.5 & 36.9 & 46.6 & 38.5 & 48.2 \\

			STCN {\cite{cheng2021rethinking}}&\checkmark& 52.0& 51.2 & 60.8 & 41.5 & 54.5  & 50.8 & 43.6 & 56.5 & 49.3 & 53.7 \\
			
			XMem {\cite{cheng2022xmem}}&\checkmark& 55.7& 54.8 & 63.3 & 45.2 & 59.7 & 53.5 & 49.5 & 62.6 & 47.1 & 54.8 \\

			AOTL {\cite{yang2021associating}}&\checkmark&  66.6& 61.4 & 71.1 & 59.4 & 74.3  & 53.8 & 50.0 & 60.3 & 47.8 & 57.1 \\
			R50\_AOTL {\cite{yang2021associating}}&\checkmark&  65.3& 61.9 & 71.4 & 56.4 & 71.6 & 54.6  & 52.9 & 63.2 & 47.5 & 54.9 \\
			SwinB\_AOTL {\cite{yang2021associating}}&\checkmark&  62.1& 58.9 & 66.5 & 54.3& 68.8 & 53.1  & 49.0 & 57.8 & 49.0 & 56.6  \\
			\textbf{Ours-Large}                     &\checkmark&\textbf{77.9}&\textbf{70.5}&\textbf{85.2}&\textbf{69.5}&\textbf{86.4} &\textbf{59.9} &\textbf{54.9} &\textbf{69.2}& \textbf{53.0} & \textbf{62.4} \\ 
		\midrule
 	{RDE}$^{*}$ {\cite{li2022recurrent}}&\checkmark& 54.3& 52.8 & 61.6 & 44.6& 58.2  & 52.2 & 44.5 & 56.0 & 49.3 & 59.1 \\  
    STCN$^{*}$ {\cite{cheng2021rethinking}}&\checkmark& 51.7 & 51.2 & 60.6 & 41.3 & 53.6 & 53.8 & 53.7 & 58.1 & 46.0 & 57.3 \\
   			XMem$^{*}$ {\cite{cheng2022xmem}}&\checkmark&  57.7& 55.6 & 64.6 & 48.6 & 61.9  & 57.9 & 51.3 & 64.5 & {53.2} & {62.7} \\ 
			\bottomrule
		\end{tabular}
	} 
	
    \label{table:PanoVOS_PanoVOS_results}
\end{table*}

\begin{figure*}[t]
	\centering
	\includegraphics[width=\textwidth]{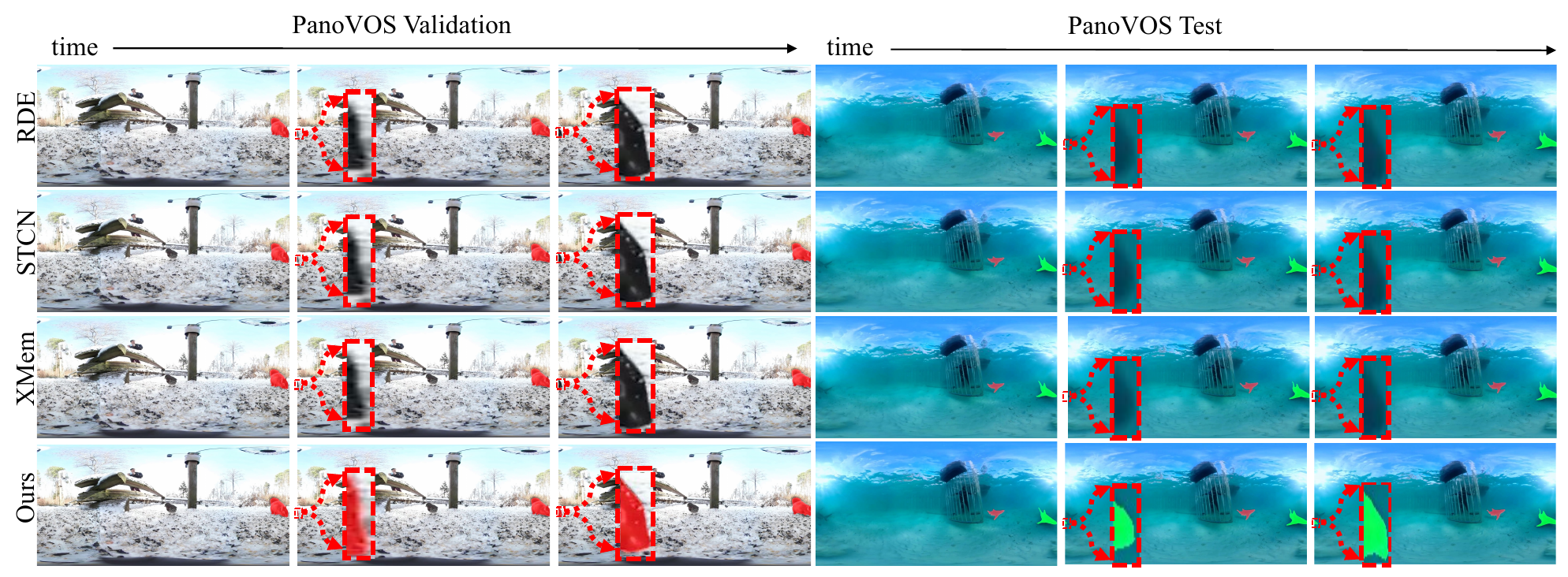} 
	\caption{\textbf{Qualitative comparison to the state-of-the-art methods}, RDE\cite{li2022recurrent}, STCN\cite{cheng2021rethinking}, and XMem\cite{cheng2022xmem}, on PanoVOS dataset. Our model performs better under the challenge of content discontinuities.  Error regions are bounded. }
\label{fig:qualitative_results}	
\end{figure*}

\section{Experiment}
In this section, we design a series of experiments to answer the following research questions related to how to tackle video object segmentation in panoramic scenes:

\noindent\textbf{RQ1}: How well are current VOS methods trained on non-panoramic videos adapted to the panoramic world?

\noindent\textbf{RQ2}: How well do variations of the foundation model Segment Anything Model~\cite{kirillov2023segment} adapt to the panoramic world?

\noindent\textbf{RQ3}: Can the proposed PanoVOS datasets bring about a consistent performance gain to VOS methods?

\noindent\textbf{RQ4}: How well does Panoramic Space Consistency Attention  contribute?

\noindent\textbf{RQ5}: What are the remained problems for  panoramic-related research?

\subsection{Implementation Details}
\noindent \textbf{Model Architecture.} We build two variants of our method with different reference bank sizes $\mathcal{R}$ for a fair comparison with previous methods. 
\textbf{Ours-Base} uses only the first frame and the previous frame as reference ($\mathcal{R}=\{1,t-1\}$), which are for the sake of high inference speed and low memory consumption.
\textbf{Ours-Large} uses multiple historical frames as reference ($\mathcal{R}=\{1+2\delta, 1+2\delta, 1+3\delta ...\}$), which follows~\cite{liu2022global, yang2021associating}. 
In our work, we set $\delta$ to 2 and 5 for training and testing respectively. For $p$ and $s$ in PSC block, we  set them as 2 and 7.

\noindent \textbf{Evaluation Metrics.} Following the standard protocol~\cite{pont20172017, perazzi2016benchmark}, we adopt the region accuracy  $\mathcal{J}$ and boundary accuracy  $\mathcal{F}$.  $\mathcal{J}$ means the Jaccard Index/Intersection over Union (IoU), which is the ratio of intersection and the joint area between predicted masks and ground truths. And $\mathcal{F}$ evaluates the accuracy of the segmentation boundary, which is computed by transforming it into a bipartite graph matching problem with predicted masks and ground truths.

\subsection{Domain Transfer Results (RQ1)}
We evaluate previous SOTA methods, which are trained on conventional datasets that are captured by pinhole cameras, on PanoVOS datasets to evaluate the domain transfer performance.  To quantify the transfer performance of advanced models trained on planar video datasets, we evaluated 15 off-the-shelf VOS models, including ~\cite{liang2020video, yang2020collaborative, yang2021collaborative, li2022recurrent, cheng2021rethinking, cheng2022xmem, yang2021associating}, and we follow official implementations and training strategies details of them.
Table~\ref{table:coco_PanoVOS_results} summarizes the domain transfer results of methods that are only trained on synthetic datasets, such as COCO~\cite{lin2014microsoft} and ECSSD~\cite{shi2015hierarchical}, on PanoVOS dataset.
Table~\ref{table:ytb_PanoVOS_results} shows the domain transfer results of state-of-the-art methods, that are trained on synthetic datasets (\eg COCO~\cite{lin2014microsoft}) and video datasets (\eg YouTube-VOS~\cite{xu2018youtube}), on our PanoVOS validation and test sets. 
By analyzing the performance of advanced VOS methods that target conventional planar videos on panoramic videos, we provide the following insights.
Firstly, the performance of current sophisticated VOS models will largely degrade when employed to tackle panoramic videos. Secondly, we can observe a trend that training on larger VOS datasets, \textit{i.e.}, YouTube-VOS~\cite{xu2018youtube} and BL30K~\cite{cheng2021rethinking} can help mitigate the gap between planar and panoramic videos.
\subsection{ Results via Visual Foundation Model (RQ2)}
To quantity the segmentation performance of different variations of the foundation model Segment Anything Model~\cite{kirillov2023segment} on PanoVOS, we evaluate the latest top performing models PerSAM~\cite{zhang2023personalize} and SAM-PT~\cite{rajivc2023segment}, as shown in Table~\ref{table:sam_PanoVOS_results}.
The performance of these models on our challenging  PanoVOS dataset is still unsatisfactory, which leaves space for further exploration.

\begin{table}[t!]
\caption{\textbf{Ablation study} of \textit{PSCAttn} module on PanoVOS.}
	\centering
	\setlength{\tabcolsep}{3.4mm}
	\resizebox{0.5\textwidth}{!}
	{
		\begin{tabular}{c|c|c|c}
			\toprule
			\multirow{2}{*}{Methods}& & \multicolumn{1}{c}{Validation}  & \multicolumn{1}{c}{Test}                \\
			\cmidrule(l){2-4} &\textit{PSCAttn} & $\mathcal{J}\&\mathcal{F}$ & $\mathcal{J}\&\mathcal{F}$ \\
			\midrule
			\multirow{2}{*}{\textbf{Ours-Base}} &  & 72.8  & 55.4  \\
		 & \checkmark & \textbf{74.0}  & \textbf{56.8} \\
		    \midrule
			\multirow{2}{*}{\textbf{Ours-Large}} &  & 74.8  & 59.5  \\
			 & \checkmark & \textbf{77.9}  & \textbf{59.9} \\
			
			\bottomrule
			
		\end{tabular}
	} 
	
    \label{table:compare_to_no_psc}
\end{table}

\begin{table}[t!]
\caption{\textbf{Comparison between our PSC attention (\textit{PSCAttn}) and cross attention (\textit{CrossAttn}) module} on PanoVOS dataset.}

	\centering
	\setlength{\tabcolsep}{3.4mm}
	\resizebox{0.5\textwidth}{!}
	{
		\begin{tabular}{c|c|c|c}
			\toprule
			\multirow{2}{*}{Methods} & Attention& \multicolumn{1}{c|}{Validation}  & \multicolumn{1}{c}{Test}    \\
			\cmidrule(l){3-4}  &Type & $\mathcal{J}\&\mathcal{F}$ & $\mathcal{J}\&\mathcal{F}$ \\
			\midrule
			\multirow{2}{*}{\textbf{Ours-Base }} & \textit{CrossAttn}   & 72.5  & 54.8  \\
			 &  \textit{PSCAttn}  & \textbf{74.0}  & \textbf{56.8} \\
		    \midrule
			\multirow{2}{*}{\textbf{Ours-Large }} & \textit{CrossAttn}  & 76.8  & 59.1  \\
		 & \textit{PSCAttn}  & \textbf{77.9}  & \textbf{59.9} \\
			
			\bottomrule
			
		\end{tabular}
	}
	
    \label{table:compare_to_crossattn}
       
\end{table}
\begin{table}[t]
  \caption{\textbf{Hyperparameter Analysis} of $p$, which enables the stitching mechanism, in PSCAttn for Ours-Large model.}
	\centering
	\setlength{\tabcolsep}{2.5mm}
	\resizebox{0.5\textwidth}{!}
	{
		\begin{tabular}{l|lllll}
			\toprule
			\multirow{2}{*}{Methods} & \multicolumn{5}{c}{PanoVOS Validation} \\
			\cmidrule(l){2-6}
			& $\mathcal{J}\&\mathcal{F}$ & $\mathcal{J}_{s}$ & $\mathcal{F}_{s}$ & $\mathcal{J}_{u}$ & $\mathcal{F}_{u}$ \\
			\midrule 
            \textit{w/o} &  73.7& 68.8 & 82.6 & 63.1 & 80.3   \\ \midrule
			$p$=$3$& 76.3 & 70.4 & 85.0 & 65.8  & 84.1 \\ 
			$p$=$5$& 74.2 & 65.3 & 79.7 & 67.5 & 84.2   \\ 
			$p$=$10$& 75.9& 68.1 & 81.9 & 68.6& 85.2   \\ 
			$p$=$15$& 75.0& 66.7 & 81.0 & 67.0 & 85.4    \\ 
            $p$=$2$ (Ours) &\textbf{77.9}&\textbf{70.5}&\textbf{85.2}&\textbf{69.5}&\textbf{86.4} \\
			\bottomrule
		\end{tabular}
	} 

	\label{table:hyper_p}

\end{table}

\subsection{Main Results on PanoVOS (RQ3)}
To evaluate the performance of previous methods on the proposed panoramic VOS dataset, we re-trained them on the training set of PanoVOS for the sake of fairness.
We report the performance in Table~\ref{table:PanoVOS_PanoVOS_results},

which demonstrates that all the previous VOS models perform worse on PanoVOS than on the traditional VOS benchmarks, \textit{e.g.}, YouTube-VOS. Our model substantially outperforms all these methods and achieves state-of-the-art on all evaluation metrics on PanoVOS, which verifies the effectiveness of our model in tackling panoramic videos.
Fig.\ref{fig:qualitative_results} visualizes some qualitative comparisons between our model and previous state-of-the-art methods on PanoVOS dataset, which shows that previous benchmarks fail to cope with content discontinuities while our model tackles them well.  

\subsection{\textcolor{black}{Ablation Study} (RQ4)}
\begin{figure}[t]
	\centering
\includegraphics[width=0.5\textwidth]{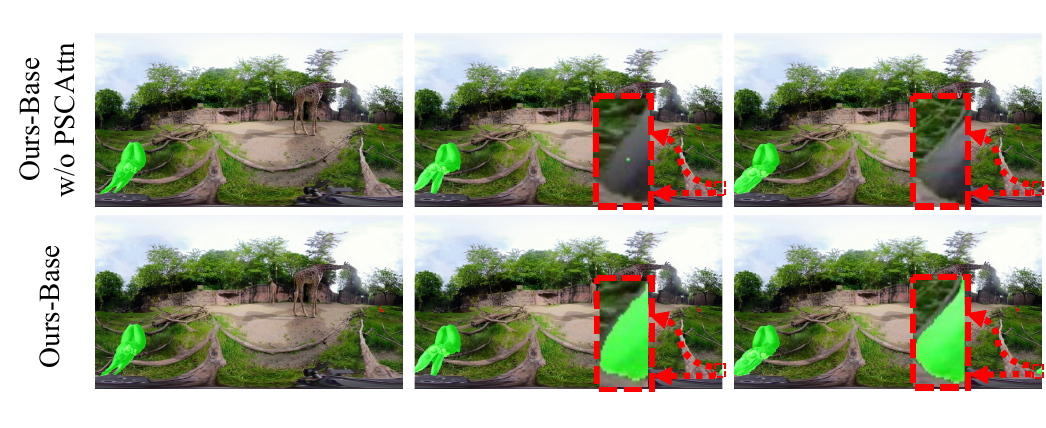}
  \caption{\textbf{Qualitative ablation study }of \textit{PSCAttn} module. }
\label{fig:ablation_results}
\end{figure}

In this section, we conduct ablation studies to demonstrate the effectiveness of the main component, \textit{i.e.}, Panoramic Space Consistency Attention (PSCAttn), of our model, with all the experiments performed based on our two model variants, \textit{i.e.}, Ours-Base and Ours-Large. For training, static image datasets are used for pre-training and PanoVOS is used for main training. Table ~\ref{table:compare_to_no_psc}  demonstrates the effectiveness of our PSCAttn module. 
Besides, Fig.~\ref{fig:ablation_results} illustrates the qualitative comparison between our default model (Ours-Base) and the setting without PSCAttn module. Our model performs better when coping with the pixel discontinuity problem. Moreover, as is shown in Table~\ref{table:compare_to_crossattn}, compared to the conventional cross-attention (CrossAttn) module, PSCAttn also achieves better performance. 
{In Table~\ref{table:hyper_p}, we analyze the hyperparameter $p$, which influences the stitching mechanism in PSCAttn, of our model ({Ours-Large}) on the PanoVOS validation set. Specifically, the highest overall performance ($\mathcal{J}\&\mathcal{F}$) is achieved when setting $p$ as 2. Compared to the setting without using the stitching mechanism ($w/o$), our model can achieve much better performance. Specifically,  our final model (Ours-Large, $p=2$) achieves more than 4\% gain in $\mathcal{J}\&\mathcal{F}$.}

\subsection{Limitation and Future Work (RQ5)}
\begin{figure}[t]
	\centering 	\includegraphics[width=0.5\textwidth]{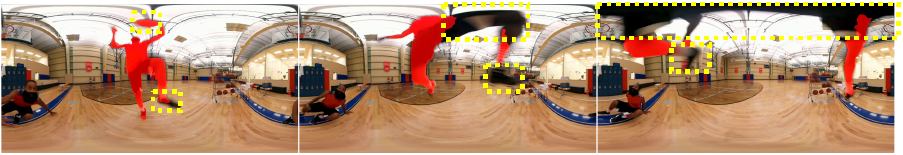} 
\caption{\textbf{Challenge.} Our model fails to segment some objects with strong distortion.}
	\label{fig:bad_case}
\end{figure}

To prompt greater progress of panoramic VOS, we also analyze the limitations of our method. Specifically, our method has no notion of severe distortion challenge since we do not employ a special design (such as deformable convolution~\cite{dai2017deformable}) to tackle deformations. That means our model may not  segment the objects with large  distortions. One such failure case is shown in Fig.~\ref{fig:bad_case}.
Besides, our panoramic dataset can be applied to broader video segmentation and tracking domains, such as referring video object segmentation~\cite{yan2024referred,li2023robust,li2023towards,li2024qdformer}, video object tracking~\cite{hong2024onetracker}, video instance segmentation~\cite{guo2023openvis}, few-shot segmentation~\cite{iqbal2022msanet}, and more broader embodied navigation tasks~\cite{xu2024customizable}. Also, it would be valuable to investigate the zero-shot segmentation performance of visual foundation models~\cite{kirillov2023segment} on our challenging panoramic dataset. We hope our work can shed light on efficient adaptation from non-panoramic to panoramic perception.


\section{Conclusion}
In this paper, we introduce a high-quality dataset, \textit{i.e.}, PanoVOS, for panoramic video object segmentation. Our PanoVOS dataset provides pixel-level instance annotations with diverse scenarios and significant motions. Based on this dataset, we evaluate 15 off-the-shelf VOS models and carefully analyze their limitations. Then, we further present our model, \textit{i.e.}, PSCFormer, which is equipped with the proposed panoramic space consistency transformer block. Our preliminary experiment demonstrates the effectiveness of our proposed model to enhance the segmentation performance and consistency in panoramic scenes. In conclusion, this provides a new challenge for video understanding, and we hope our PanoVOS dataset can attract more researchers to pay attention to panoramic videos.

\section*{Acknowledgements}
This work was supported  in part by National Natural Science Foundation of China (No.62072112), and Scientific and Technological Innovation Action Plan of Shanghai Science and Technology Committee (No.22511101502, No.22511102202 and No.21DZ2203300).

%
%
\bibliographystyle{splncs04}
\bibliography{main}
\end{document}